# A Maximum Likelihood Approach For Selecting Sets of Alternatives

**Ariel D. Procaccia**
Computer Science Department
Carnegie Mellon University
arielpro@cs.cmu.edu

**Sashank J. Reddi**
Machine Learning Department
Carnegie Mellon University
sjakkamr@cs.cmu.edu

**Nisarg Shah**
Computer Science Department
Carnegie Mellon University
nkshah@cs.cmu.edu

## Abstract

We consider the problem of selecting a subset of alternatives given noisy evaluations of the relative strength of different alternatives. We wish to select a $k$-subset (for a given $k$) that provides a maximum likelihood estimate for one of several objectives, e.g., containing the strongest alternative. Although this problem is $\mathcal{NP}$-hard, we show that when the noise level is sufficiently high, intuitive methods provide the optimal solution. We thus generalize classical results about singling out one alternative and identifying the hidden ranking of alternatives by strength. Extensive experiments show that our methods perform well in practical settings.

## 1 Introduction

The initial motivation for this paper stemmed from discussions with the inventors of EteRNA (http://eterna.cmu.edu). Similarly to Foldit [9], EteRNA is a scientific discovery game where the goal is to design RNA molecules that fold into stable structures. Thousands of human players propose RNA designs weekly, but only a relatively small number $k$ of them can be synthesized in a laboratory to discover whether they actually fold well. To choose $k$ designs to synthesize, players vote over the proposed designs using a voting rule known as $k$-*approval*: each player reports up to $k$ favorite designs, each of which is awarded one point. The $k$ designs that receive the most points are synthesized, and the best design is identified.

The aggregation of opinions via $k$-approval has several shortcomings. For example, players typically do not consider all proposed designs, and in particular designs that receive votes early in the voting process gain visibility and therefore usually accumulate more votes, leading to a snowball effect. One alternative is to elicit players' opinions about pairs of designs, making sure that each proposed design is compared with at least several others. Regardless of how players' opinions are elicited, one would need to answer the question: *what is our objective in aggregating opinions?* For EteRNA, the answer is simple; since biochemists are seeking a stable design for a specific molecule, but will only use a single proposed design, we would like to select a set of $k$ designs that includes at least one great design. Including one great design is sufficient, as it will be singled out when the $k$ selected designs are synthesized.

Of course, this setup is not confined to the realm of human computation; it also arises naturally, for example, in the process of *new product development (NPD)*. A product can have many different potential designs. A $k$-subset of these designs is selected based on a market survey, and (possibly costly) prototypes are manufactured for the selected designs. The prototypes are then evaluated by a focus group that definitively singles out the best design. Again, the goal is to include at least one great design among the prototypes that are evaluated.

A natural approach views elicited opinions over alternatives (designs, in the examples) as noisy estimates of a true, hidden ranking of the alternatives in terms of quality. Given such a noise model, the examples presented above call for the selection of the subset of alternatives that is most likely to contain the top alternative of the true ranking, that is, the truly strongest alternative. However, other settings may require the selection of a subset of alternatives that is most likely to possess a different property. In this paper we study this maximum likelihood estimation (MLE) framework under several objectives and noise models.

### 1.1 Our model and results

Our model consists of two components: the noise model and the objective function. We are interested in two basic noise models (see, e.g, [4]) that govern how the dataset is obtained given a hidden true ranking. The first model—the *noisy comparisons* model—corresponds to independent evaluations of pairs of alternatives, where each evaluation is consistent with the true ranking with a fixed probability $p \in (1/2, 1)$. For the second model—the *noisy orders* model (also known as the *Mallows* model [14])—we imagine voters submitting complete rankings over the alternatives. The probability of observing a ranking is exponentially small in its Kendall Tau distance to the true ranking, i.e., the number of pairs of alternatives on which the two rankings disagree. The second model is consistent with EteRNA's current voting method, where in theory players are expected to consider all designs, but it is much more natural when the number of alternatives (e.g., product designs) is small. Our positive results hold with respect to a general noise model—the *noisy choice model*—that includes both basic models as special cases.

For the second component, we focus on three objective functions. Objective 1 is the one discussed above: select a subset that maximizes the probability of including the top alternative of the true ranking. Objective 2 aims to select a $k$-subset of alternatives that coincides with the top $k$ alternatives of the true ranking; this objective is natural, for example, when choosing a team of workers to carry out a task that requires multiple workers with identical skills (here the alternatives are the workers). Objective 3 seeks to select an ordered tuple of alternatives of length $k$ that maximizes the probability of coinciding with the $k$-prefix of the true ranking. In other words, we are trying to single out the $k$ top alternatives as before, but in the correct order; this objective is closely aligned with web search applications. Note that the three objectives coincide when $k = 1$. We consider Objective 1 to be the most natural and important among the three, and indeed our exposition concentrates on this objective.

We prove that computing the optimal solution under all three objectives is $\mathcal{NP}$-hard for any nontrivial value of $k$ (for Objectives 1 and 2, the case of $k = m$, where $m$ is the number of alternatives, is trivial). However, our analytical results indicate that the optimal solutions for special cases correspond to intuitive (and sometimes tractable) methods. In particular, our analytical results focus on the case where the level of noise is very high. There are two compelling reasons for considering such noisy settings. First, if the level of noise is not very high, any reasonable method would be able to single out the true ranking with relatively little data and high confidence, and therefore maximizing our objectives becomes a nonissue. Second, as we discuss below, our experiments clearly indicate that methods that perform well in theory with respect to a very high noise level also perform well in practice.

For Objective 1, we introduce the *extended scoring method* to single out a $k$-subset of alternatives. We prove that under the noisy choice model, when the noise level is sufficiently high, any optimal solution is in the solution space provided by the extended scoring method.[1] Interestingly, under noisy orders the extended scoring method reduces to a well-known voting rule (which maps a vector of rankings submitted by voters to a selected alternative) called *Borda count*, and its special case for noisy comparisons also provides a highly intuitive and tractable method. To our surprise, it turns out that the extended scoring method also yields the optimal solution for Objective 2 when the noise level is sufficiently high.

For Objective 3 we present the *scored tuples* method, and prove that it gives the optimal solution when the noise level is high. Interestingly, under the noisy orders model, this method coincides with Borda count when $k = 1$, and with another famous rule known as the *Kemeny* rule when $k = m$. Intermediate values of $k$ give a sequence of optimal voting rules that connects Borda count with Kemeny; we believe that this insight is of independent interest to social choice theory.

Finally, we conduct extensive experiments that compare the extended scoring method (for Objectives 1 and 2) and the scored tuples method (for Objective 3) with other methods. Our experiments indicate that the proposed methods, which are theoretically optimal under high noise, also outperform other methods under practical noise levels. Moreover, in cases where we are able to compute the MLE (i.e., the optimal solution), its performance and that of the proposed methods are almost indistinguishable.

### 1.2 Related work

Young [18] studied maximum likelihood estimators under (a variation of) the noisy orders model and two objectives: identifying the top alternative of the true ranking (which coincides with our objectives for $k = 1$), and identifying the MLE ranking. In fact, Young's paper is based on work done by the marquis de Condorcet roughly two centuries earlier [10]. Young found that when the noise level is sufficiently high, the winner according to Borda count is the MLE for the top alternative, and regardless of the noise level the Kemeny rule is an MLE for the true ranking. Our main results for Objectives 1 and 2 generalize Young's results for Borda count by extending them to a more powerful noise model and (more importantly) to dif-ferent values of $k$. Our result for Objective 3 connects Young's results for Borda ($k = 1$) and Kemeny ($k = m$) via a sequence of optimal voting rules for intermediate values of $k$, while again generalizing the noise model.

---

[1]Subtleties with respect to tie breaking are discussed in detail in Section 3.

A series of relatively recent AI papers deal with voting rules as MLEs (see, e.g., [6, 8, 17, 16]). In particular, in a UAI 2005 paper, Conitzer and Sandholm [6] reverse Young's question by asking for which common voting rules there exist (constrained) noise models such that the voting rules are MLEs for the top alternative or the true ranking. One section of the paper of Xia and Conitzer [16] studies a setting where a set of alternatives must be selected, under a noise model that is different from ours, where there is a set of winners of size $k$ but no underlying ranking. They provide a single result for this setting: an evaluation problem that is related to identifying the set of winners is $\mathcal{NP}$-hard.

Under standard noise models such as noisy orders and noisy comparisons, computing the MLE true ranking is also $\mathcal{NP}$-hard [3, 5], but there is a significant amount of work on this problem (see, e.g., [1, 2, 7, 12, 4]). For example, Braverman and Mossel [4] provide polynomial time algorithms that compute the MLE ranking with high probability. However, we see below that selecting the top $k$ elements of the MLE ranking is not the optimal solution to our objectives (with the obvious exception of Objective 3 for $k = m$) and moreover our experiments show that this method provides poor performance with respect to our objectives in practice.

## 2 The Model

We denote $[k] = \{1, \ldots, k\}$. In addition, let $\arg\max_{s \in S}^k H(s)$ be the set of $k$-subsets of $S$ (let $|S| = t$) with largest values under the given function $H$. In other words, for each order $(s_1, \ldots, s_t)$ of the elements of $S$ such that $H(s_i) \geq H(s_{i+1})$ for all $i \in [t-1]$, $\arg\max_{s \in S}^k H(s)$ includes the set $\{s_1, \ldots, s_k\}$.

We consider a set of alternatives $A$; denote $|A| = m$. We will use small letters to denote specific alternatives. Let $L(A)$ be the set of permutations (which we think of as linear orders or rankings) on $A$, where each permutation is a bijection $\sigma : A \to \{1, 2, \ldots, m\}$. Hence, $\sigma(a)$ denotes the position of alternative $a$ in $\sigma$, and $\sigma^{-1}(i)$ denotes the alternative at the $i$th position, i.e., the $i$th most preferred alternative when $\sigma$ is viewed as a ranking over the alternatives. In particular, $\sigma(a) < \sigma(b)$ denotes that $a$ is preferred to $b$ under $\sigma$. We let $\sigma^{-1}([k]) = (\sigma^{-1}(1), \ldots, \sigma^{-1}(k))$ denote the ordered tuple consisting of the $k$-prefix of $\sigma$.

We assume that there exists a true hidden order $\sigma^* \in L(A)$ over the alternatives, which reflects their true strengths. We also make the standard assumption that $\sigma^*$ is selected using a uniform prior over $L(A)$. Let $a^* = (\sigma^*)^{-1}(1)$ denote the best alternative under $\sigma^*$.

Our objective is to find a "good" set of alternatives given noisy observations. We consider two standard noise models (see, e.g., [4]).

### 2.1 Noisy comparisons and tournaments

In the *noisy comparisons* model, a pairwise preference $a \succ b$ denotes that alternative $a$ is preferred to alternative $b$. We imagine that each pair of alternatives is presented to $n$ voters (with possibly different sets of voters for each pair). The preferences returned by the voters are independently consistent with the true ranking $\sigma^*$ with a fixed probability $1/2 < p < 1$. Hence, the dataset $D$ is a set of comparisons where each pair of alternatives appears exactly $n$ times, for a fixed value of $n$. Note that the case of $n = 1$ with relatively high value of $p$ can also represent the aggregate opinion of many voters.

We think of the dataset $D$ as corresponding to a slightly nonstandard *weighted tournament*. Denote by $n_{ab}$ the number of $a \succ b$ votes. The tournament $T_D$ is a directed graph where the vertex set is the set of alternatives, there are edges between each pair of alternatives in both directions, and the weight of the edge $e = (a, b)$ is $w_e = n_{ab}$.[2]

### 2.2 Noisy orders and ranked voting

In our second model, a fixed set of $n$ voters provide rankings over all alternatives. In this *noisy orders model* (also known as the *Condorcet noise model* [10]), each ranking is generated independently by drawing pairwise preferences similarly to the noisy comparisons model, except that the process is restarted if the generated vote has a cycle (e.g. $a \succ b \succ c \succ a$). Concisely, the probability of drawing each ranking $\sigma_i$ given the true order $\sigma^* = \sigma$ is proportional to $p^{\binom{m}{2} - d_K(\sigma_i, \sigma)} \cdot (1-p)^{d_K(\sigma_i, \sigma)}$. The distance $d_K(\cdot, \cdot)$ is the *Kendall tau* distance between two rankings, which counts their number of disagreements on pairs of alternatives. Probabilities are normalized using a normalization constant which can be shown to be independent of the true ranking $\sigma^*$ (see, e.g., [13]). Note that this model is equivalent to the *Mallows* model [14], which is widely used in machine learning and statistics.

A *voting rule* (also known as a *social choice function*) is a function $f : L(A)^n \to A$ that accepts $n$ rankings $(\sigma_1, \sigma_2, \cdots, \sigma_n) \in L(A)^n$ as input and outputs a selected alternative. We will informally use the same term to refer to functions that output a ranking of the alternative, i.e., an element of $L(A)$; such functions are also known as *social welfare functions*.

---

[2]Typically tournaments have exactly one directed edge between each pair of vertices.

Below we consider a number of well-known voting rules; we begin with two that will play a special role. Under *Borda count* each voter $i$ awards $m - \sigma_i(a)$ points to each alternative $a \in A$, and an alternative with most points overall is selected. The *Kemeny* rule selects a ranking $\pi \in L(A)$ that minimizes $\sum_{i=1}^n d_K(\pi, \sigma_i)$. Informally, the Kemeny ranking minimizes the number of disagreements with voters over pairs of alternatives.

In Section 6 we consider some additional voting rules as benchmarks. Under *Maximin*, the score of an alternative $a$ is $\min_{a' \in A \setminus \{a\}} |\{i \in [n] : \sigma_i(a) < \sigma_i(a')\}|$. Under *plurality*, each voter $i$ awards one point to $\sigma_i^{-1}(1)$, and under *k-approval*, one point to each of $\sigma_i^{-1}(1), \ldots, \sigma_i^{-1}(k)$.

### 2.3 A generalization: The noisy choice model

We next present the *noisy choice model*, a general framework that unifies both models; to the best of our knowledge this model is novel. The model is characterized by two properties. First, a notion of $n_{ab}$ for all alternatives $a \in A$ and $b \in A \setminus \{b\}$ that denotes the degree to which $a$ is preferred over $b$ in the input. For a fixed $n$ and all $a \in A$ and $b \in A \setminus \{a\}$, $n_{ab} + n_{ba} = n$.

Second, a likelihood of a dataset $D$ given that the true order is $\sigma^* = \sigma$,

$$\Pr[D|\sigma^* = \sigma] = \frac{\gamma^{d(\sigma, D)}}{Z_\gamma}, \quad (1)$$

where the normalizing constant $Z_\gamma$ is independent of $\sigma$ (although it might depend on the parameters of the model), and the distance function is defined as $d(\sigma, D) = \sum_{a,b \in A: \sigma(a) < \sigma(b)} n_{ba}$. The distance intuitively measures the amount of disagreement between the ranking and the dataset on pairs of alternatives. Crucially, the likelihood of the dataset diminishes exponentially as its distance from the true order increases.

The noisy comparisons model and noisy orders model both fall under this more general framework. Indeed, the notion of $n_{ab}$ is clear in both models: it is simply the number of votes that prefer $a$ to $b$. It is also easy to verify that $\gamma = (1-p)/p$ for both models.

In addition to noisy comparisons and noisy orders, the noisy choice model can capture other practical models of inputs. For example, consider the model where inputs are noisy partial orders of a fixed length $l$ generated as follows. For each voter, $l$ alternatives are chosen uniformly at random from the set of all alternatives. Then a partial order is generated according to the Mallows Model over the chosen $l$ alternatives. Alternatively, we can consider a model where each voter reports an unweighted tournament over the alternatives, i.e., individual preferences may be "irrational".

Note that $\gamma$ is a measure of the level of noise; $\gamma \approx 0$ induces a distribution that is highly centered around the true order and $\gamma \approx 1$ induces a distribution that is close to the uniform distribution. We also remark that the distance function $d(\cdot, \cdot)$ under the noisy orders model is just the sum of Kendall tau distances of a particular ranking from the input orders.

## 3 Including the Top Alternative

We first consider the case where we want to select a $k$-subset of alternatives that is most likely to contain $a^*$, the top alternative of the true ranking $\sigma^*$.

**Objective 1** *Given $k$, find a $k$-subset of alternatives that maximizes the probability of containing the best alternative, i.e., a subset in*

$$\arg\max_{S \subseteq A, |S|=k} \Pr[a^* \in S|D].$$

A crucial observation regarding Objective 1 is that

$$\Pr[a^* \in S|D] = \sum_{a \in S} \Pr[a^* = a|D] \propto \sum_{a \in S} \Pr[D|a^* = a],$$

where the last step follows from Bayes' rule and the assumption of uniform prior over rankings. The following observation is immediately implied.

**Observation 3.1** *The optimal solution to Objective 1 under the noisy choice model is any subset in*

$$\arg\max_{a \in A}^k \Pr[a^* = a|D].$$

In words, we choose the $k$ most likely alternatives according to their probabilities of coinciding with the top alternative of the true ranking. Equivalently, one could select any subset in $\arg\max_{a \in A}^k \Pr[D|a^* = a]$ since we have assumed a uniform prior over rankings. Despite Observation 3.1, finding an optimal solution to Objective 1 is computationally hard.

**Theorem 3.2** *For any $k \in [m-1]$, computing an optimal solution to Objective 1 is $\mathcal{NP}$-hard under noisy orders and noisy comparisons.*

The theorem's proof appears in the full version of the paper.[3] Note that the theorem also proves $\mathcal{NP}$-hardness under the noisy choice model because noisy orders and noisy comparisons are just special cases. The intuition behind the proof is as follows. Young [18] demonstrated (via an example) that when $p$ close to 1, under a specific noise model that is very similar to the noisy comparisons model, the optimal solution with respect to Objective 1 with $k = 1$ coincides with the first element of an MLE of the true ranking $\sigma^*$. We show that this result holds under both noisy comparisons and noisy orders and also prove that computing the first element of an MLE ranking is $\mathcal{NP}$-hard by utilizing the $\mathcal{NP}$-hardness of computing the MLE ranking itself. Here, an MLE ranking is a minimum feedback ranking under noisy comparisons (see, e.g., [4]), and a Kemeny ranking under noisy orders [18], which are both $\mathcal{NP}$-hard to compute [5, 3]. Finally, we leverage the case of $k = 1$ to extend the $\mathcal{NP}$-hardness to other values of $k$.

---
[3]The full version of the paper is available from: http://www.cs.cmu.edu/~arielpro/papers.html.

Because of these computational connections, it is natural to wonder whether the optimal solution to Objective 1 is given by taking the top $k$ elements of the MLE ranking for any $k$ and any value of $p$. However, Young [18] also showed that this is not the case when $p$ is close to $1/2$ even for $k = 1$ (in fact he did not consider the case of $k > 1$). Indeed, he showed that when $p$ is close to $1/2$, in his example the MLE ranking $\sigma$ is the only ranking that puts the alternative $\sigma(1)$ first and has significant probability, whereas a different alternative (the winner under Borda count) appears first in rankings that individually have smaller probability than the MLE ranking, but combined have a larger overall probability. Below we extend this result by showing that when $p$ is sufficiently close to $1/2$, the problem is indeed tractable for any value of $k$ under our general noisy choice model, and in fact the solution coincides with intuitive methods. Although the theoretical guarantees are for the case where $p$ is close to $1/2$ (i.e., very noisy settings, $\gamma \approx 1$ in Equation (1)), the experiments in Section 6 show that the methods developed here also work well when the noise level is lower.

Our general method, which we refer to as the *extended scoring method*,[4] is well defined for input data generated according to the noisy choice model. For an alternative $a \in A$, let the score of $a$ be

$$\mathrm{sc}(a) = \sum_{b \in A \setminus \{a\}} n_{ab}. \qquad (2)$$

We choose the top $k$ alternatives according to their score, i.e., we return a set in $\arg\max_a^k \mathrm{sc}(a)$.

**Theorem 3.3** *For every $n$ and $m$ there exists $\gamma' < 1$ such that for all $\gamma \geq \gamma'$, the optimal solutions to Objective 1 under the noisy choice model are in $\arg\max_a^k \mathrm{sc}(a)$.*

The theorem's formulation leaves open the possibility that some sets in $\arg\max_a^k \mathrm{sc}(a)$ are far from being optimal. However, the theorem's proof in fact shows that for any $\delta > 0$ there is sufficiently large $\gamma$ such that *every* $S \in \arg\max_a^k \mathrm{sc}(a)$ is optimal up to $\delta$, i.e., for any $S' \subseteq A$ such that $|S'| = k$,

$$\Pr[a^* \in S | D] \geq \Pr[a^* \in S' | D] - \delta.$$

**Proof of Theorem 3.3** Let $\gamma = 1 - \epsilon$, $\epsilon > 0$. By Observation 3.1, we know that the optimal solution is given by $\arg\max_a^k \Pr[D | a^* = a]$. In our model

$$\Pr[D | a^* = a] = \sum_{\sigma \in L(A), \sigma^{-1}(1) = a} \Pr[D | \sigma^* = \sigma]$$
$$= \sum_{\sigma \in L(A), \sigma^{-1}(1) = a} \frac{\gamma^{d(\sigma, D)}}{Z_\gamma}.$$

Let $L_a(A) = \{\sigma \in L(A) | \sigma^{-1}(1) = a\}$. Thus $|L_a(A)| = (m-1)!$. Define an objective function $f(a) = Z_\gamma \cdot \Pr[D | a^* = a]$. Since $Z_\gamma$ is a constant, the optimal solution is also given by $\arg\max_a^k f(a)$. Using $\gamma = 1 - \epsilon$ and the fact that $(1-\epsilon)^t \geq 1 - t \cdot \epsilon$ for any $t \in \mathbb{N}$, we obtain the following lower bound on our objective function $f$:

$$f(a) \geq \hat{f}(a) = \sum_{\sigma \in L_a(A)} (1 - \epsilon \cdot d(\sigma, D)). \qquad (3)$$

In addition, the gap between $f$ and $\hat{f}$ can be upper bounded using

$$|(1-\epsilon)^t - (1 - t \cdot \epsilon)| \leq \sum_{i=2}^{t} \binom{t}{i} \cdot \epsilon^i \leq 2^t \cdot \epsilon^2$$

for $\epsilon < 1$. It follows that for every $a \in A$,

$$\begin{aligned} f(a) - \hat{f}(a) &\leq \sum_{\sigma \in L_a(A)} 2^{d(\sigma, D)} \cdot \epsilon^2 \\ &\leq \epsilon^2 \cdot (m-1)! \cdot 2^{n \cdot \binom{m}{2}}. \end{aligned} \qquad (4)$$

The theorem will now follow by proving two statements: $\arg\max_a^k f(a) \subseteq \arg\max_a^k \hat{f}(a)$, and $\arg\max_a^k \hat{f}(a) = \arg\max_a^k \mathrm{sc}(a)$. We use the following claim.

**Claim 3.4** *For every $a \in A$,*

$$\hat{f}(a) = C_\epsilon + \epsilon \cdot (m-1)! \cdot \mathrm{sc}(a)$$

*where $C_\epsilon$ depends only on $\epsilon$ (and not on $a$).*

---
[4]We use the term "extended" to avoid confusion with generalized scoring rules [15].

**Proof** First, we simplify $\hat{f}$ by summing the individual terms in Equation (3).

$$\hat{f}(a) = (m-1)! - \epsilon \cdot \sum_{\sigma \in L_a(A)} d(\sigma, D). \quad (5)$$

Furthermore, note that $\sum_{\sigma \in L_a(A)} d(\sigma, D)$ equals

$$\sum_{\sigma \in L_a(A)} \sum_{x \in A, y \in A \setminus \{x\}} n_{yx} \cdot \mathbb{1}[\sigma(x) < \sigma(y)]$$

$$= \sum_{\sigma \in L_a(A)} \sum_{x \in A, y \in A \setminus \{x\}} (n - n_{xy}) \cdot \mathbb{1}[\sigma(x) < \sigma(y)]$$

$$= n \cdot \sum_{\sigma \in L_a(A)} \sum_{x \in A, y \in A \setminus \{x\}} \mathbb{1}[\sigma(x) < \sigma(y)]$$

$$- \sum_{x \in A, y \in A \setminus \{x\}} n_{xy} \cdot \sum_{\sigma \in L_a(A)} \mathbb{1}[\sigma(x) < \sigma(y)]$$

The symbol $\mathbb{1}$ represents the indicator function. For the first term it holds that

$$n \cdot \sum_{\sigma \in L_a(A)} \sum_{x \in A, y \in A \setminus \{x\}} \mathbb{1}[\sigma(x) < \sigma(y)]$$

$$= n \cdot \sum_{\sigma \in L_a(A)} \binom{m}{2} = n \cdot (m-1)! \cdot \binom{m}{2}.$$

To analyze the second term we consider three cases separately: (i) $x = a$, $y \in A \setminus \{a\}$, (ii) $x \in A \setminus \{a\}$, $y = a$, and (iii) $x \in A \setminus \{a\}$, $y \in A \setminus \{a, x\}$. If $x = a$, then for every $y \in A \setminus \{a\}$ and every $\sigma \in L_a(A)$ it holds that $a \succ y$, and hence $n_{ay}$ is multiplied by $(m-1)!$. Similarly, if $y = a$, then for every $x \in A \setminus \{a\}$, $n_{xa}$ is multiplied by 0. If $x \in A \setminus \{a\}$ and $y \in A \setminus \{a, x\}$, then $n_{xy}$ is multiplied by $(m-1)!/2$ because $x \succ y$ in exactly half of the rankings in the summation. Thus the second terms equals

$$(m-1)! \cdot \left( \sum_{y \in A \setminus \{a\}} n_{ay} + \frac{1}{2} \cdot \sum_{x \in A \setminus \{a\}, y \in A \setminus \{a, x\}} n_{xy} \right)$$

$$= (m-1)! \cdot \text{sc}(a) + \frac{(m-1)!}{2} \cdot n \cdot \binom{m-1}{2},$$

since $n_{xy} + n_{yx} = n$ for every $x \neq y$. Combining both terms and substituting back,

$$\sum_{\sigma \in L_a(A)} d(\sigma, D)$$

$$= n \cdot (m-1)! \cdot \binom{m}{2}$$

$$- \frac{(m-1)!}{2} \cdot n \cdot \binom{m-1}{2} - (m-1)! \cdot \text{sc}(a)$$

$$= C'_\epsilon - (m-1)! \cdot \text{sc}(a),$$

for some constant $C'_\epsilon$ independent of $a$. Plugging this into Equation (5) we get

$$\hat{f}(a) = C_\epsilon + \epsilon \cdot (m-1)! \cdot \text{sc}(a)$$

as required. ■(Claim 3.4)

Claim 3.4 directly implies that $\arg\max_a^k \hat{f}(a) = \arg\max_a^k \text{sc}(a)$. It therefore remains to prove that $\arg\max_a^k f(a) \subseteq \arg\max_a^k \hat{f}(a)$. For this purpose it is sufficient to show that for every $a, a' \in A$, $\hat{f}(a) > \hat{f}(a')$ implies $f(a) > f(a')$.

Note that $\hat{f}(a) > \hat{f}(a')$ implies $\text{sc}(a) \geq \text{sc}(a') + 1$ (since scores are integers) and using Claim 3.4 this implies $\hat{f}(a) \geq \hat{f}(a') + \epsilon \cdot (m-1)!$. Therefore

$$f(a) \geq \hat{f}(a) \geq \hat{f}(a') + \epsilon \cdot (m-1)!$$

$$\geq f(a') - \epsilon^2 \cdot (m-1)! \cdot 2^{n \cdot \binom{m}{2}} + \epsilon \cdot (m-1)!.$$

The third transition follows from Equation (4). Setting $\epsilon < 2^{-n \cdot \binom{m}{2}}$, i.e., $\gamma > \gamma' = 1 - 2^{-n \cdot \binom{m}{2}}$, we have that $f(a) > f(a')$ as required. ■

The notion of score underlying the extended scoring method (Equation (2)) provides an intuitive reflection of the quality of an alternative. In the special case of the noisy comparisons model, $\text{sc}(a) = \sum_{x \in A \setminus \{a\}} n_{ax}$ is just the sum of weights of the outgoing edges from $a$ in the weighted tournament defined in Section 2. Hence, the extended scoring method reduces to picking $k$ alternatives with highest weighted outdegrees.

In the special case of the noisy orders model,

$$\text{sc}(a) = \sum_{x \in A \setminus \{a\}} n_{ax} = \sum_{x \in A \setminus \{a\}} \sum_{i=1}^n \mathbb{1}[\sigma_i(a) < \sigma_i(x)]$$

$$= \sum_{i=1}^n \sum_{x \in A \setminus \{a\}} \mathbb{1}[\sigma_i(a) < \sigma_i(x)] = \sum_{i=1}^n (m - \sigma_i(a)),$$

which is exactly the Borda score of alternative $a$. Hence, the extended scoring method reduces to picking $k$ alternatives with maximum Borda scores. Theorem 3.3 thus extends Young's result for Borda count [18] from the noisy orders model to the noisy choice model and from $k = 1$ to any value of $k$ under Objective 1.

## 4 Identifying the Top Subset

Under Objective 1 we choose $k$ alternatives, each of which is likely to be the top alternative $a^*$. However, in principle it may be the case that each of these alternatives is either the top-ranked alternative or among the bottom-ranked alternatives. In this section we seek to identify the set of top $k$ alternatives, but we will see that the solution in fact coincides with the solution to Objective 1.

**Objective 2** *Given $k$, find the $k$-subset of alternatives that maximizes the probability of coinciding with the top $k$ alternatives of the true hidden order, i.e., a subset in*

$$\underset{S \subseteq A, |S|=k}{\arg\max} \ \Pr[S = \{(\sigma^*)^{-1}(i)\}_{i \in [k]} | D].$$

It is easy to see that this objective coincides with Objective 1 for $k = 1$, and hence it is $\mathcal{NP}$-hard for $k = 1$ under noisy orders and noisy comparisons. As before, we can extend this observation to any $k \in [m-1]$ for all three models (the proof appears in the full version of the paper).

Next, we show that the extended scoring method is again optimal when $\gamma \approx 1$.

**Theorem 4.1** *For every $n$ and $m$ there exists $\gamma' < 1$ such that for all $\gamma \geq \gamma'$, the optimal solutions to Objective 2 under the noisy choice model are in $\arg\max_a^k sc(a)$.*

Theorem 4.1 (whose proof is given in the full version of the paper) suggests that the alternatives selected by the extended scoring method are not only good candidates for the best alternative individually, but as a whole they also make a good "team", i.e, when put together they are the most likely to coincide with the top $k$ alternatives. Theorems 4.1 and 3.3 together provide an argument in favor of the extended scoring method and, in particular, they strongly advocate Borda count when inputs are noisy orders.

## 5 Identifying the Top Tuple

In this section we study an objective that is even more ambitious than Objective 2: not only do we want to correctly identify the top $k$ alternatives of $\sigma^*$, we seek to identify them *in the correct order*.

**Objective 3** *Given $k$, find the ordered $k$-tuple that maximizes the probability of coinciding with the $k$-prefix of the true hidden order, i.e., a tuple in*

$$\underset{(a_1, a_2, \ldots, a_k) \in A^k}{\arg\max} \ \Pr[(\sigma^*)^{-1}([k]) = (a_1, a_2, \ldots, a_k) | D].$$

Objective 3 coincides with its predecessors when $k = 1$, and reduces to finding an MLE for the true ranking when $k = m$. In fact we are able to prove that computing the optimal solution to Objective 3 is $\mathcal{NP}$-hard for any $k \in [m]$ under noisy orders and noisy comparisons (the proof appears in the full version of the paper).

To tackle Objective 3 we propose a new method, which we call the *scored tuples* method. Similarly to the extended scoring method, it maximizes a lower bound of the stated objective function and provides optimality guarantees when $\gamma \approx 1$. We first extend the definition of the noisy choice model's distance function to compute the distance between a $k$-tuple $(a_1, a_2, \ldots, a_k)$ and a dataset $D$; this distance is defined as

$$d((a_1, a_2, \ldots, a_k), D) = \sum_{1 \leq i < j \leq k} n_{a_j a_i}.$$

Next, for a $k$-tuple $(a_1, a_2, \ldots, a_k)$, define the score of the tuple as

$$sc(a_1, a_2, \ldots, a_k) = \sum_{i=1}^{k} sc(a_i) - d((a_1, a_2, \ldots, a_k), D), \quad (6)$$

where $sc(a_i)$ is defined as in Equation (2). We select a $k$-tuple in $\arg\max_{(a_1, a_2, \ldots, a_k) \in A^k} sc(a_1, a_2, \ldots, a_k)$.

It may seem that overloading the notation of $sc(a)$ as we have (via Equation (2) and Equation (6) with a tuple of length 1) may create inconsistencies, but in fact the $k = 1$ case of Equation (6) reduces to Equation (2).

It turns out that the above method provides guarantees with respect to Objective 3 that are equivalent to those that we were able to obtain for previous objectives.

**Theorem 5.1** *For every $n$ and $m$ there exists $\gamma' < 1$ such that for all $\gamma \geq \gamma'$, the optimal solutions to Objective 3 under the noisy choice model are in $\arg\max_{(a_1, a_2, \ldots, a_k) \in A^k} sc(a_1, a_2, \ldots, a_k)$.*

The proof of Theorem 5.1 is given in the full version of the paper; let us consider its implications with respect to the noisy orders model. Since maximizing $sc(a_1, a_2, \ldots, a_k)$ is optimal when $\gamma \approx 1$, it must be the case (and indeed it is easy to verify) that this solution reduces to finding the Borda winner when $k = 1$ and reduces to finding the Kemeny ranking when $k = m$. Thus the optimal solution for $k = 1, \ldots, m$ induces a range of voting rules where Borda count lies on one extreme, and the Kemeny rule lies on the other. We find it intriguing (and of independent interest to social choice theory) that Borda count and Kemeny are connected via a sequence of "optimal" voting rules.

We remark that the Kemeny rule has been suggested as a method of aggregating the search results given by different search engines; in particular it can be formally argued that spam websites will appear at the bottom of the aggregate ranking [11]. However, since users are usually only interested in the top 50 results or so, our results suggest that the scored tuples method with a relatively small $k$ may outperform the Kemeny rule (same method with $k=m$) in this context.

Finally, an important note is that although computing $\arg\max_{(a_1,a_2,\ldots,a_k)\in A^k} \text{sc}(a_1, a_2, \ldots, a_k)$ is $\mathcal{NP}$-hard in general, it can be computed in polynomial time for constant $k$. In some practical settings one would only need to select a constant number of alternatives (consider, e.g., the search engine example given above) and the rule can therefore be easily applied.

## 6 Experiments

We performed simulations under the noisy comparisons model and the noisy orders model. We experimented with various values of the accuracy parameter $p$ (recall that $\gamma = (1-p)/p$). Although the theoretical guarantees hold only for $p$ close to $1/2$ (which corresponds to $\gamma$ close to 1), we observed that for various values of $p$ in the practical range the methods suggested in this paper significantly outperform various other known methods.

The graphs present results for particular values of $p$ for which the probabilities are visible; the probabilities quickly go to 1 for higher values of $p$ and go to 0 for lower values of $p$. The results are also verified empirically for various values of $m$ (number of alternatives) and $n$ (number of voters) and $p$ (accuracy parameter) under both models. Error bars correspond to 95% confidence intervals and thus verify that the suggested methods show statistically significant improvement over other methods. In all the graphs shown below each point is computed by taking the average over 10000 iterations.

We remark that the simulations are symmetric with respect to the true order, i.e., it does not matter which true order we begin with as long as the methods do not use any information about it. If we perform simulations on a fixed true order, it becomes necessary to break ties uniformly at random, which is what we do.

For noisy orders we compared Borda count for Objectives 1 and 2, and the scored tuples method for Objective 3, against the Kemeny rule (which is the MLE ranking), $k$-approval, plurality, maximin, and maximizing unweighted outdegree in the tournament where there is an edge from $a$ to $b$ if a majority of voters rank $a$ above $b$ (a.k.a. *Copeland*). For noisy comparisons we compared weighted outdegree for Objectives 1 and 2, and the scored tuples method for Objective 3, against the Minimum Feedback ranking (which is the MLE ranking), unweighted outdegree, and Maximin.

For Objective 1, we were also able to estimate the actual optimal solution using Observation 3.1. For this we needed to sample rankings according to $\Pr[\sigma^* = \sigma|D]$. One obstacle is that it can be shown that even computing $\Pr[\sigma^* = \sigma|D]$ is $\mathcal{NP}$-hard. We used the fact that $\Pr[\sigma^* = \sigma|D] \propto \Pr[D|\sigma^* = \sigma] \propto \gamma^{d(\sigma,D)}$ and sampled rankings using the Metropolis-Hastings algorithm. Note though that for larger values of the parameters the optimal solution is very difficult to estimate, whereas finding a Borda winner is a trivial computational task.

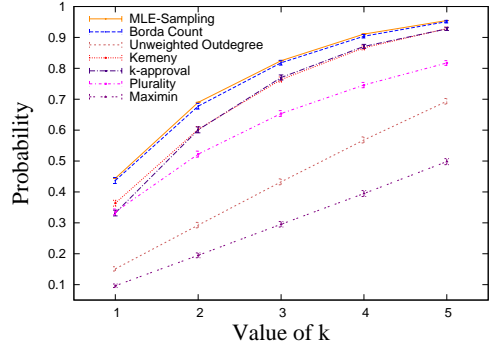

Fig. 1: Objective 1, Noisy Orders, m=10, n=10, p=0.55.

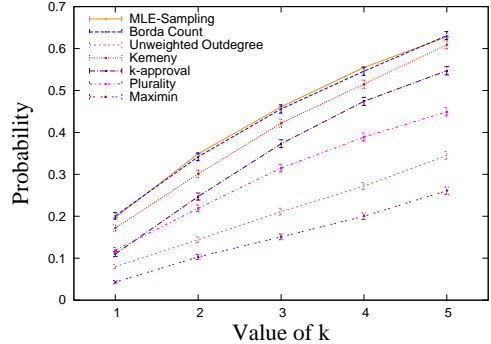

Fig. 2: Objective 1, Noisy Orders, m=20, n=100, p=0.505.

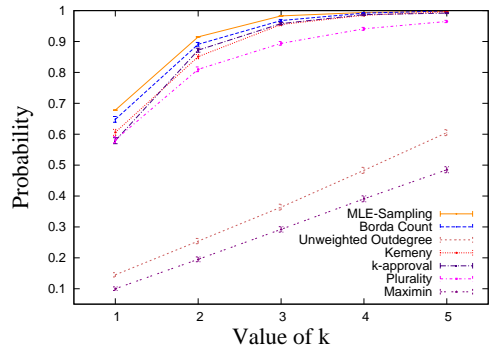

Fig. 3: Objective 1, Noisy Orders, m=10, n=10, p=0.6.

Figure 1 shows simulations for 10 alternatives, 10 voters and $p = 0.55$ under noisy orders. Clearly Borda count outperforms other methods by a large margin, and in fact it is almost indistinguishable from the actual optimal solution. Extensive simulations show that similar results hold for a large number of alternatives and/or a large number of voters as well. An example with 20 alternatives, 100 voters and $p = 0.505$ under noisy orders is shown in Figure 2.

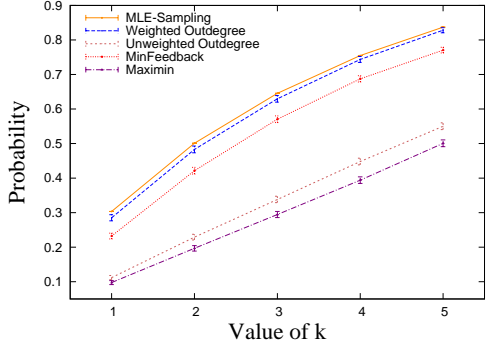

Fig. 4: Objective 1, Noisy Comparisons, m=10, n=10, p=0.55.

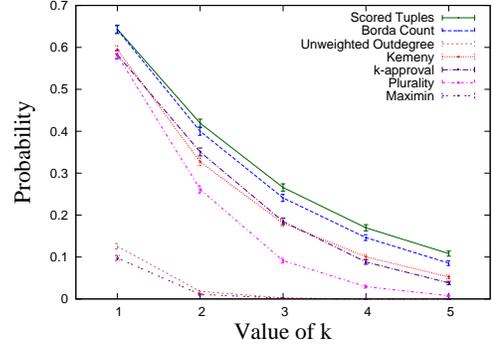

Fig. 6: Objective 3, Noisy Orders, m=10, n=10, p=0.55.

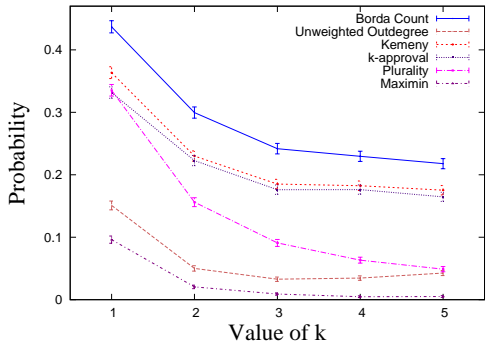

Fig. 5: Objective 2, Noisy Orders, m=10, n=10, p=0.55.

As mentioned above, the value of $p$ is chosen such that the graphs span the probability spectrum (probabilities are not constantly 1 or 0) but similar results hold for other values of $p$ as well. For example, Figure 4 shows that Borda count dominates other methods when $p = 0.6$. However, the margin is smaller. Indeed, for this relatively high value of $p$, most methods under consideration achieve excellent accuracy with only ten voters.

We also performed detailed simulations under the noisy comparisons model and observed similar results. Figure 4 shows a sample simulation for 10 alternatives, 10 voters and $p = 0.55$. In this case weighted outdegree (which is the special case of the extended scoring method) outperforms other methods by a statistically significant margin.

Under Objective 2, the extended scoring method again outperforms other methods as shown in Figure 5 (with 10 alternatives, 10 voters and $p = 0.55$ under noisy orders). As expected, under Objective 3 the scored tuples method outperforms other methods. Figure 6 shows simulations with 10 alternatives, 10 voters and $p = 0.55$ under noisy orders. We do not provide additional graphs for noisy comparisons due to lack of space, but the results are similar: the methods that are theoretically optimal for $p$ close to $1/2$ outperform other methods for practical values of $p$.

## 7 Discussion

We began our exposition with the human computation angle, and we would like to revisit it now that we have gained some new insights. Voting is a natural and almost ubiquitous tool in human computation systems. However, the designers of these systems usually employ simplistic voting rules such as plurality or $k$-approval (as EteRNA does). Our results suggest that the choice of voting rule can significantly affect performance. For example, given that we are indeed interested in singling out one great design, switching from $k$-approval to Borda count in EteRNA can provide significant benefits. Of course we cannot expect players to rank all the proposed designs, but we can work with partial rankings or pairwise comparisons (as described in Section 2). We find it exciting that social choice theory can help improve human computation systems. Indeed, it is difficult to apply the principles of social choice (e.g., voting rules as MLEs) to political elections, because it is almost impossible to switch voting rules. In contrast, human computation provides a perfect testbed for these principles.

# References

[1] N. Ailon, M. Charikar, and A. Newman. Aggregating inconsistent information: Ranking and clustering. In *Proceedings of the 37th Annual ACM Symposium on Theory of Computing (STOC)*, pages 684–693, 2005.

[2] N. Alon. Ranking tournaments. *SIAM Journal of Discrete Mathematics*, 20(1–2):137–142, 2006.

[3] J. Bartholdi, C. A. Tovey, and M. A. Trick. Voting schemes for which it can be difficult to tell who won the election. *Social Choice and Welfare*, 6:157–165, 1989.

[4] M. Braverman and E. Mossel. Noisy sorting without resampling. In *Proceedings of the 19th Annual ACM-SIAM Symposium on Discrete Algorithms (SODA)*, pages 268–276, 2008.

[5] P. Charbit, S. Thomassé, and A. Yeo. The minimum feedback arc set problem is NP-hard for tournaments. *Combinatorics, Probability & Computing*, 16(1):1–4, 2007.

[6] V. Conitzer and T. Sandholm. Common voting rules as maximum likelihood estimators. In *Proceedings of the 21st Annual Conference on Uncertainty in Artificial Intelligence (UAI)*, pages 145–152, 2005.

[7] V. Conitzer, A. Davenport, and H. Kalagnanam. Improved bounds for computing Kemeny rankings. In *Proceedings of the 21st AAAI Conference on Artificial Intelligence (AAAI)*, pages 620–626, 2006.

[8] V. Conitzer, M. Rognlie, and L. Xia. Preference functions that score rankings and maximum likelihood estimation. In *Proceedings of the 21st International Joint Conference on Artificial Intelligence (IJCAI)*, pages 109–115, 2009.

[9] S. Cooper, F. Khatib, A. Treuille, J. Barbero, J. Lee, M. Beenen, A. Leaver-Fay, D. Baker, and Z. Popović. Predicting protein structures with a multiplayer online game. *Nature*, 466:756–760, 2010.

[10] M. de Condorcet. Essai sur l'application de l'analyse à la probabilité de décisions rendues à la pluralité de voix. Imprimerie Royal, 1785. Facsimile published in 1972 by Chelsea Publishing Company, New York.

[11] C. Dwork, R. Kumar, M. Naor, and D. Sivakumar. Rank aggregation methods for the web. In *Proceedings of the 10th International World Wide Web Conference (WWW)*, pages 613–622, 2001.

[12] C. Kenyon-Mathieu and W. Schudy. How to rank with few errors. In *Proceedings of the 39th Annual ACM Symposium on Theory of Computing (STOC)*, pages 95–103, 2007.

[13] T. Lu and C. Boutilier. Learning Mallows models with pairwise preferences. In *Proceedings of the 28th International Conference on Machine Learning (ICML)*, pages 145–152, 2011.

[14] C. L. Mallows. Non-null ranking models. *Biometrika*, 44:114–130, 1957.

[15] L. Xia and V. Conitzer. Generalized scoring rules and the frequency of coalitional manipulability. In *Proceedings of the 9th ACM Conference on Electronic Commerce (EC)*, pages 109–118, 2008.

[16] L. Xia and V. Conitzer. A maximum likelihood approach towards aggregating partial orders. In *Proceedings of the 22nd International Joint Conference on Artificial Intelligence (IJCAI)*, pages 446–451, 2011.

[17] L. Xia, V. Conitzer, and J. Lang. Aggregating preferences in multi-issue domains by using maximum likelihood estimators. In *Proceedings of the 9th International Joint Conference on Autonomous Agents and Multi-Agent Systems (AAMAS)*, pages 399–408, 2010.

[18] H. P. Young. Condorcet's theory of voting. *The American Political Science Review*, 82(4):1231–1244, 1988.